\documentclass[10pt, a4paper]{article}
\usepackage{multirow}
\usepackage{booktabs}

\usepackage[final]{lrec2026} % this is the new style
\title{COACH meets QUORUM: A Framework and Pipeline for Aligning User, Expert and Developer Perspectives in LLM-generated Health Counselling}

\name{
Yee Man Ng$^{\ast}$, Bram van Dijk$^{\dagger}$, Pieter Beynen$^{\diamond}$, Otto Boekesteijn$^{\diamond}$,\\
{\bf \large Joris Jansen$^{\diamond}$, Gerard van Oortmerssen$^{\ast}$, Max van Duijn$^{\ast}$, Marco Spruit$^{\ast\dagger}$}
}

\address{$^{\ast}$Leiden University, \{y.m.ng, m.r.spruit, g.van.oortmerssen, m.j.van.duijn\}@liacs.leidenuniv.nl \\
         $^{\dagger}$Leiden University Medical Center, b.m.a.van\_dijk@lumc.nl\\
         $^{\diamond}$Healthy Chronos, \{pieter.beynen, otto.boekesteijn, joris.jansen\}@healthychronos.com \\
         }

\abstract{
Systems that collect data on sleep, mood, and activities can provide valuable lifestyle counselling to populations affected by chronic disease and its consequences. Such systems are however challenging to develop; besides \textit{reliably} extracting patterns from user-specific data, systems should also contextualise these patterns with validated medical knowledge to ensure the \textit{quality} of counselling, and generate counselling that is \textit{relevant} to a real user. We present QUORUM, a new evaluation framework that unifies these developer-, expert-, and user-centric perspectives, and show with a real case study that it meaningfully tracks convergence and divergence in stakeholder perspectives. We also present COACH, a Large Language Model-driven pipeline to generate personalised lifestyle counselling for our \texttt{Healthy Chronos} use case, a diary app for cancer patients and survivors. Applying our framework shows that overall, users, medical experts, and developers converge on the opinion that the generated counselling is relevant, of good quality, and reliable. However, stakeholders also diverge on the tone of the counselling, sensitivity to errors in pattern-extraction, and potential hallucinations. These findings highlight the importance of multi-stakeholder evaluation for consumer health language technologies and illustrate how a unified evaluation framework can support trustworthy, patient-centered NLP systems in real-world settings. 
\\ \newline  }
\begin{document}

\maketitleabstract

\section{Introduction}
People living with the consequences of cancer (treatment) often suffer from long-term effects such as fatigue and mental health issues \cite{akechi1999fatigue}. While it is known that healthy lifestyle changes like increasing physical activity positively impact health outcomes and quality of life, for both cancer patients and survivors \cite{hoedjes2022psychosocial}, they currently experience the state of lifestyle counselling in healthcare as not readily available, too fragmented, and general \cite{vanAken_2025,Drbohlav_2025,Tuinman_2024}. This demonstrates the need for personalised lifestyle counselling, through reliable pattern extraction from user-specific health records and with validated sources, in order to produce high-quality and evidence-based answers to consumer-health questions. 
%reliable pattern extraction from user-specific health records, as well as contextualisation of these patterns, with validated sources %to produce high-quality counselling and insights that matter to users. 
%to provide high-quality, evidence-based answers to consumer-health questions that are meaningful to users.
While many AI-driven systems that generate counselling exist, their evaluation methodologies are often fragmented and lack realistic contexts \cite{lai2025evaluation,raji2021ai}.

This is why this paper contributes \textbf{QUORUM} (\textbf{QU}ality, \textbf{O}utcome \textbf{R}eliability, and \textbf{U}ser-relevance from \textbf{M}ultiple stakeholders), a new evaluation framework that aligns with the goals and interests of three key types of stakeholders in the development of NLP systems for health. From a \textit{user-centric} perspective, counselling is \textit{relevant} to users if the advice aligns with their situation, increases the likelihood of action based on the counselling, and if they are satisfied with the tone and length of the counselling. \textit{Experts} on the other hand (in our case cancer information specialists from \url{kanker.nl}, henceforth `experts') focus on the \textit{quality} of the counselling with respect to its correctness, and the appropriate tone and length for the population and disease they know well. \textit{Developers/researchers} (two authors, henceforth `developers') have as vantage point \textit{reliability}, i.e., baseline performance on key metrics such as faithfulness and completeness of the counselling, and hallucination rates. We argue that these three perspectives must be included in the evaluation of AI-systems that support specific medical populations. Since we cannot expect such stakeholders to evaluate the output of AI systems by the same standards, they provide irreducible, complementary perspectives that only jointly indicate whether a given system is likely to have positive, real-world impact.

\begin{figure}[t]
    \centering
    \includegraphics[width=.95\columnwidth]{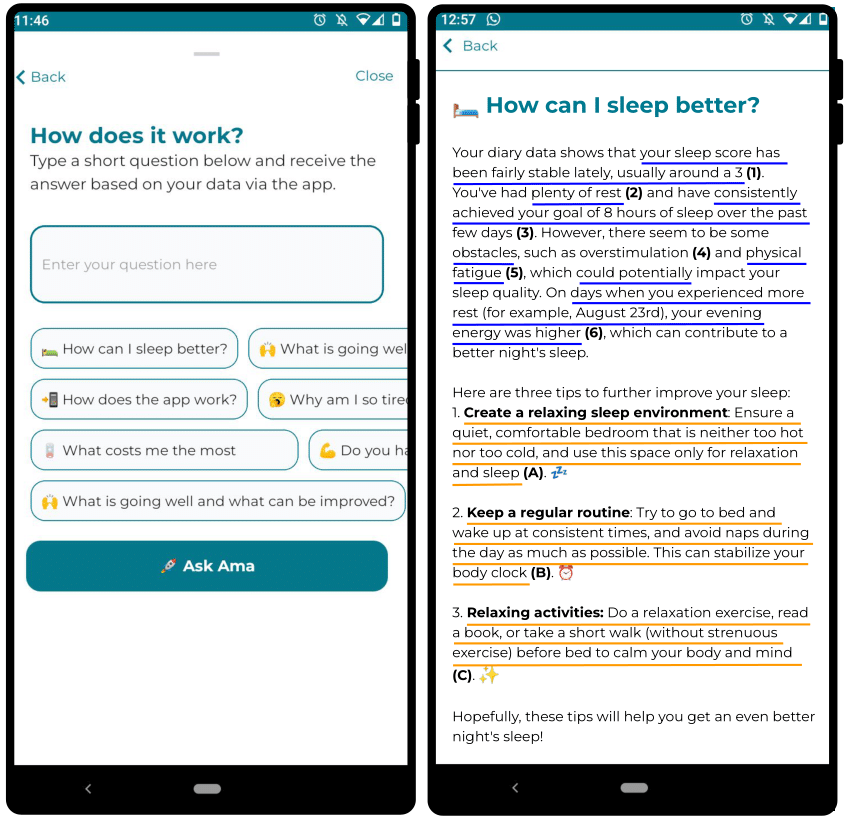}
    \caption{Left: user interface for submitting queries. Right: generated counselling for query `How can I sleep better?' Blue underlining illustrate \textit{claims about the user data}; orange indicates \textit{contextualisation statements} that contextualise found patterns with information from the knowledge database. Claims and statements are labelled with (1), (A), etc. but this is hidden from the user.}
    \label{fig:hc_ux_response_question}
\end{figure}

We show QUORUM in action with a concrete use case: the Dutch \texttt{Healthy Chronos} application. This is a digital diary app that aids cancer patients and survivors by tracking sleep, activities, mood, and custom user-defined goals over time. We develop the Large Language Model (LLM) driven pipeline \textbf{COACH} (\textbf{C}ontextualised \textbf{O}utcome-\textbf{A}daptive \textbf{C}ounselling for \textbf{H}ealth) for \texttt{Healthy Chronos}, which uses Retrieval Augmented Generation (RAG) \cite{rag_2020} to generate personalised lifestyle counselling. The pipeline ingests a user query about their data (e.g. \textit{How can I sleep better?}), after which COACH retrieves relevant chunks for that query from the external, validated knowledge database \texttt{kanker.nl}. COACH then contextualises the patterns extrapolated from the user data with the retrieved chunks into personalised counselling (for an impression see \autoref{fig:hc_ux_response_question}). 
By using QUORUM on COACH, we demonstrate that relevant, good quality, and reliable lifestyle counselling is possible with LLMs. Nineteen Dutch app users acknowledge that the counselling aligns with their situation, they intend to follow up on the counselling, and endorse its tone and length. Six independent experts from \texttt{kanker.nl} deem the counselling of good quality in terms of the correctness of the contextualisation of extracted user patterns, and its tone and length. Technical evaluation by two developers shows COACH is reliable, as most claims (79\%) are consistent with user data; moreover, for virtually all claims (97\%), the relevant information has been retrieved. Still, we also find that 22\% of the generated counselling statements are not strictly traceable to the knowledge database, thus are hallucinations (though not of a harmful kind). 

Besides this convergence, QUORUM also highlights divergence between stakeholders. For example, users and experts differ on the desired tone of the response, and differ with developers in the sensitivity to errors and hallucination in the reported patterns and counselling. These results highlight the utility of QUORUM and COACH for other populations and medical conditions, by helping building systems with a larger real-world impact.

\section{Background}
Consulting the internet for information on lifestyle change to improve one's health and quality of life is increasingly common  \cite{raeside2022navigating}. Still, people struggle with finding information fitting their specific queries, due to answer complexity, potentially incorrect information, and the lack of alignment with their situation \cite{bondarenko_2021,pugachev_2023,kayser2015enhancing}. Though LLMs are recognised as potentially useful for providing lifestyle counselling due to their language and reasoning abilities, a recent overview shows that particularly evaluation methodologies for LLM-driven lifestyle counselling are fragmented, weak, and lack realistic contexts \cite{lai2025evaluation}. 

Related work has further shown that for LLM-driven chatbots in health, stakeholders can have diverging perspectives relevant to system development. For example, whereas users may prioritize the tone of the interaction, professional caregivers may emphasize safety and clinical validation aspects \cite{van2025welzijn}. Still, current work typically targets only a subset of stakeholders, like relevance from a user perspective but no further quality assessment of system output \cite{meywirth2024designing}, or expert evaluation without including real users \cite{merrill2024transforming}. While such studies address important issues, often with novel AI techniques, extrapolating findings to real populations remains hard.

However, a study close to our setup is \citet{steenstra2024virtual}, and it concerns an LLM-driven tool for Motivational Interviewing regarding alcohol use. It also adopts a three-way evaluation with different stakeholders, where humans evaluated the tone of the LLM-generated outputs, researchers evaluated the presence of appropriate MI techniques in the responses, and independent experts overall tested the system by role-playing interactions. Yet, \citeauthor{steenstra2024virtual}'s work differs from ours in that their tool is not embedded in the real world. 

Regarding system development, research has shown that many LLM-driven systems for lifestyle counselling still rely on proprietary models and cloud services \cite{lai2025evaluation}. For startups this is perhaps understandable, as self-hosting LLMs is costly, while cloud access to capable LLMs is cheap \cite{nandagopaltokens}; these are key considerations for startups like our \texttt{Healthy Chronos} use case that inform some methodological choices we make below.

\begin{table*}[t]
\centering
\scriptsize

\setlength{\tabcolsep}{4pt}
\renewcommand{\arraystretch}{1.15}

\begin{tabular}{
@{\extracolsep{\fill}}
>{\raggedright\arraybackslash}p{60pt}   % Perspective
>{\raggedright\arraybackslash}p{120pt}  % Statement
>{\raggedright\arraybackslash}p{50pt}   % Variable
>{\raggedright\arraybackslash}p{70pt}   % Measurement
>{\raggedright\arraybackslash}p{120pt}  % Explanation
}
\toprule
\textbf{Perspective} (N) & \textbf{Statement evaluated} & \textbf{Variable} & \textbf{Measurement} & \textbf{Interpretation} \\
\toprule

\multirow{6}{=}{Users (19)}
& \multirow{2}{*}{1 \textit{The response aligns with my situation}}
& \multirow{2}{*}{Alignment}
& \multirow{2}{*}{5-point Likert scale}
& Avg. score indicating alignment with user's situation (mood, energy, etc.) \\

& \multirow{2}{*}{2 \textit{I will follow the tips in the response}}
& \multirow{2}{*}{Follow-up}
& \multirow{2}{*}{5-point Likert scale}
& Avg. score indicating whether user is likely to act on the response \\

& 3 \textit{The response has the right tone}
& Tone
& 5-point Likert scale
& Avg. score of response tone quality  \\

& 4 \textit{The response has the right length}
& Length
& 5-point Likert scale
& Avg. score of response length \\

\midrule

\multirow{3}{=}{Experts (6)}
& 1 \textit{The contextualisation is correct}
& Correctness
& 5-point Likert scale
& Avg. score of contextualisation quality \\

& 2 \textit{The response has the right tone}
& Tone
& 5-point Likert scale
& Avg. score of response tone quality  \\

& 3 \textit{The response has the right length}
& Length
& 5-point Likert scale
& Avg. score of response length  \\

\midrule

\multirow{5}{=}{Developers (2)}
& 1 \textit{This claim is consistent with the data}
& Faithfulness
& Binary
& \% of claims supported by user data \\

& 2 \textit{For this claim the relevant user variables were retrieved}
& \multirow{2}{*}{Completeness}
& \multirow{2}{*}{Binary}
& \% of claims for which relevant columns were retrieved from user data \\

& 3 \textit{This contextualisation statement can be traced to the relevant chunks found by the pipeline}
& \multirow{2}{*}{Hallucination}
& \multirow{2}{*}{Binary}
& \% of claims grounded in the knowledge database (\texttt{kanker.nl})  \\
\bottomrule
\end{tabular}
\caption{Contents (translated from Dutch) of  the QUORUM evaluation framework. Five-point Likert scales are structured as \textit{strongly disagree }(1)\textit{, disagree }(2)\textit{, neutral }(3)\textit{, agree }(4)\textit{, strongly agree }(5) .}
\label{tab:eval_framework}
\end{table*}

\section{Methodology}

%\subsection{Prompt engineering}
%LLM-driven systems for lifestyle counselling have to analyse user-specific datasets robustly and reliably. Here, we describe various insights from the field sometimes dubbed `tabular understanding' that were relevant to and incorporated in our pipeline, particularly in the way we set up prompting and further inputs. 
%First, regarding order, we separated instructions and examples from `external' information, i.e. table representation and chunks from the knowledge database, as it is known that keeping external information at the end improves performance; also, in the same work LLMs were found to be sensitive to different ways to serialize tabular inputs \cite{sui_table_2024}, which is why in initial experiments we explored different formats like JSON and Markdown (but found Markdown to work best). Second, we know that in-context learning provides an efficient way to improve performance without parameter updates \cite{fang2024large}, hence we described in the prompt for three recurrent topics in user queries (physical fatigue, mental fatigue, daily activities) the relevant columns in the user data, to improve the capacity of the LLM to find the right column but also prime it to consider relations between columns. Third, given that we know that headers provide crucial information for LLMs to improve understanding column contents and referencing to other columns \cite{singha2023tabular}, we also included descriptions of key column names and their data types, for example, "headers containing 'goal' always indicate booleans".

\subsection{QUORUM Framework}
QUORUM unifies three complementary perspectives from the literature. First, the \textbf{model-centric perspective} that emphasises well-known standardized metrics (e.g., recall) on common tasks (e.g., retrieving relevant elements). These metrics are important as they allow quickly probing a system's base performance and comparison with other systems, but they are poor proxies for a system's real-world impact \cite{wang2024understanding,li-etal-2025-llms-cant}. This is nevertheless a common perspective to adopt for AI \textbf{developers}, and in our framework it involves faithfulness, completeness, and hallucination metrics as common to RAG evaluation \cite{papageorgiou2025evaluating}. Faithfulness concerns the extent to which generated claims about the user data are consistent with that user's data; completeness concerns the extent to which all relevant information in the user data for some query was retrieved; hallucination concerns the existence of contextualisation statements in the counselling that cannot be traced to the knowledge database. (Recall the difference between \textit{claims about user data} and \textit{contextualisation statements} from \autoref{fig:hc_ux_response_question}.)

The \textbf{user-centric perspective}, on the other hand, focuses on how well model behaviour actually helps a \textbf{real user} to reach a specific goal, thus is about the relevance of a system's output \cite{wang-etal-2024-user}. We asked users to evaluate the alignment of the response with their personal situation, the tone and length of the response, and the likelihood the user would follow up on the counselling provided. These more subjective measures aim to capture the relevance of the response to the user.

We also incorporate the \textbf{expect-centric perspective}. In a health context, besides user and developer, \textbf{medical experts} as stakeholders often have different concerns \cite{van2025welzijn}, for example about the overall adequacy of the medical information in the response, because actions may be based on it, and because users may re-use the generated information in real medical consultations. Thus, we ask experts to evaluate the correctness of the contextualisation, based on their expertise on the (consequences of) cancer, as well as their opinion on the style and length of the response. 

\begin{figure*}[t]
    \centering
    \includegraphics[width=\linewidth]{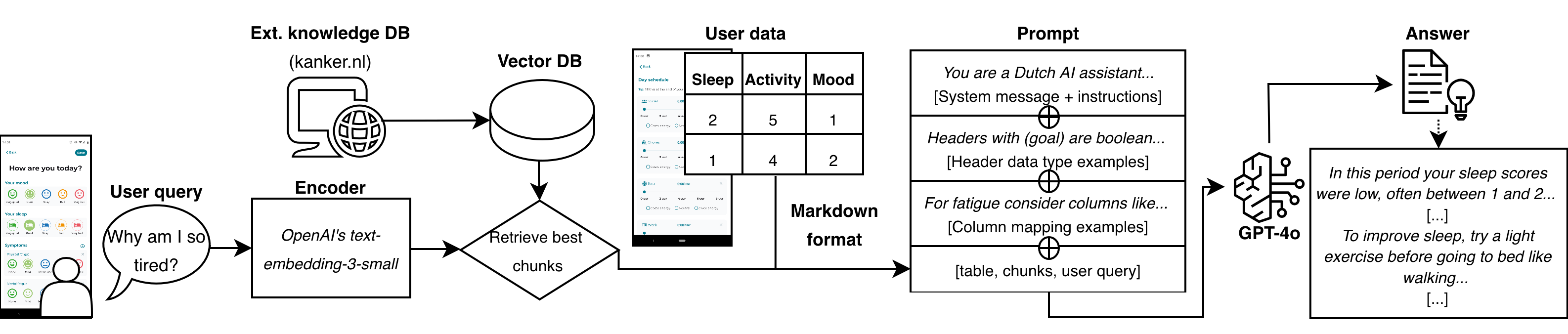}
    \caption{Overview of the pipeline flow and inputs.}
    \label{fig:pipeline_overview}
\end{figure*}

\subsubsection{Holistic, iterative evaluation}
In this way we aim to incorporate a holistic perspective on the generated counselling, that strikes a good balance between technical (model-centric), subjective (user-centric), and medical (expert-centric) aspects.  Combining these three perspectives is vital, as single perspectives cannot by themselves provide a good estimation of the real-world impact of AI systems in health; they are complementary and one perspective cannot be reduced to another. Though they do overlap in topics, the perspectives try to capture those opinions that can be reasonably expected from different stakeholders given their background, goals and interests. 

For example, users could be less likely to verify reported patterns or lifestyle advice, if an answer is in line with what they already believe; this is a common issue in dealing with the fluent output of LLMs nowadays and known as confirmation and consistency bias \cite{nirman2025fool}. Or, they could be less likely to do so due to automation bias: an overreliance on the capabilities of algorithms in generating valid information \cite{boonprakong2025quantifying}. This shows how model- and expert-centric evaluation can complement each other. On the other hand, an AI system with perfect reliability may still have little real-world impact because few of its counselling may be and meaningful to a user; here the user-centric perspective provides the best proxy for real impact. What is still missing then is some assurance that the reliable and relevant output, is also of good quality from an expert-centric perspective; only experts can adequately assess medical content.     

QUORUM was conceived of as a quick probe in the context of \textit{iterative} system development, where especially for interactive systems, many brief stakeholder evaluations need to be done to incorporate their views in future versions of the system  \cite{van2011holistic,van2025welzijn}. After evaluation is done with each group, a QUORUM user can compute per perspective a grand average that indicates how well the system at issue is doing from that perspective, while the individual variables provide details on the perspective's key aspects. See \ref{tab:eval_framework} for a schematic overview of variables in QUORUM; we further show and discuss outputs of our framework in \autoref{sec:results}. 

\subsubsection{Data collection}
Here we describe the procedures for data collection from users, experts, and researchers. All protocols were approved by the Leiden University Ethics Review Board. Though we share the source code of our pipeline in the Supplementary Materials (and later publicly online), we cannot share user data itself nor \texttt{Healthy Chronos} source code due to privacy and intellectual property considerations.

\paragraph{Users} Nineteen Dutch users were via the app recruited to participate in this study. As shown in \autoref{fig:hc_ux_response_question}, users can query their data, but the app also suggests query topics based on frequency. In our evaluation, users evaluate responses on the question: \textit{What is going well and what can be improved?} One reason for this is that this query covers a large part (41\%) of a random sample of 100 queries. Also, this query has a broad scope, meaning that the pipeline is forced to correctly retrieve and analyse potentially many different relevant variables, which can be more challenging compared to more specific queries (e.g. about sleep). Users received the query and generated response in the app, and then completed a short survey on the statements given in \autoref{tab:eval_framework}. They were asked to evaluate the full response, and were also invited to submit open remarks for each query-response pair. This amounted to nineteen evaluated query-response pairs. We did not collect further demographic data due to privacy considerations, but the user base of \texttt{Healthy Chronos} contains mostly Dutch adolescents and young adults.

\paragraph{Experts} Six experts from \texttt{kanker.nl} consented to evaluating in total 50 random user query-response pairs, and they evaluated for each pair the four statements given in \autoref{tab:eval_framework}. Experts were instructed to focus for correctness on the \textit{contextualisation}, (see \autoref{fig:hc_ux_response_question}, as due to privacy concerns it was not feasible to share user data with them, and because the contextualisation is based on the knowledge database (\texttt{kanker.nl} they were familiar with. Besides providing scores, experts were also invited to submit remarks at every query-response pair.

\paragraph{Developers} Two developers evaluated in two rounds in total 66 \textit{claims about the user data} and 31 \textit{contextualisation statements} (see \autoref{fig:hc_ux_response_question}) from 10 random query-response pairs. Statements 1 and 2 in \autoref{tab:eval_framework} were used for the claims about user data, whereas for the contextualisation statement 3 was used. After the developers initially evaluated the whole set independently, disagreements were discussed to consensus afterwards. 

We computed measures for interrater reliability for experts and developers, for various variables, and therefore describe these metrics in tandem with our results in \autoref{sec:results}.  

\subsection{Pipeline}\label{sec:methodolody:pipeline}
\begin{figure}[t]
    \centering
    \includegraphics[width=0.95\columnwidth]{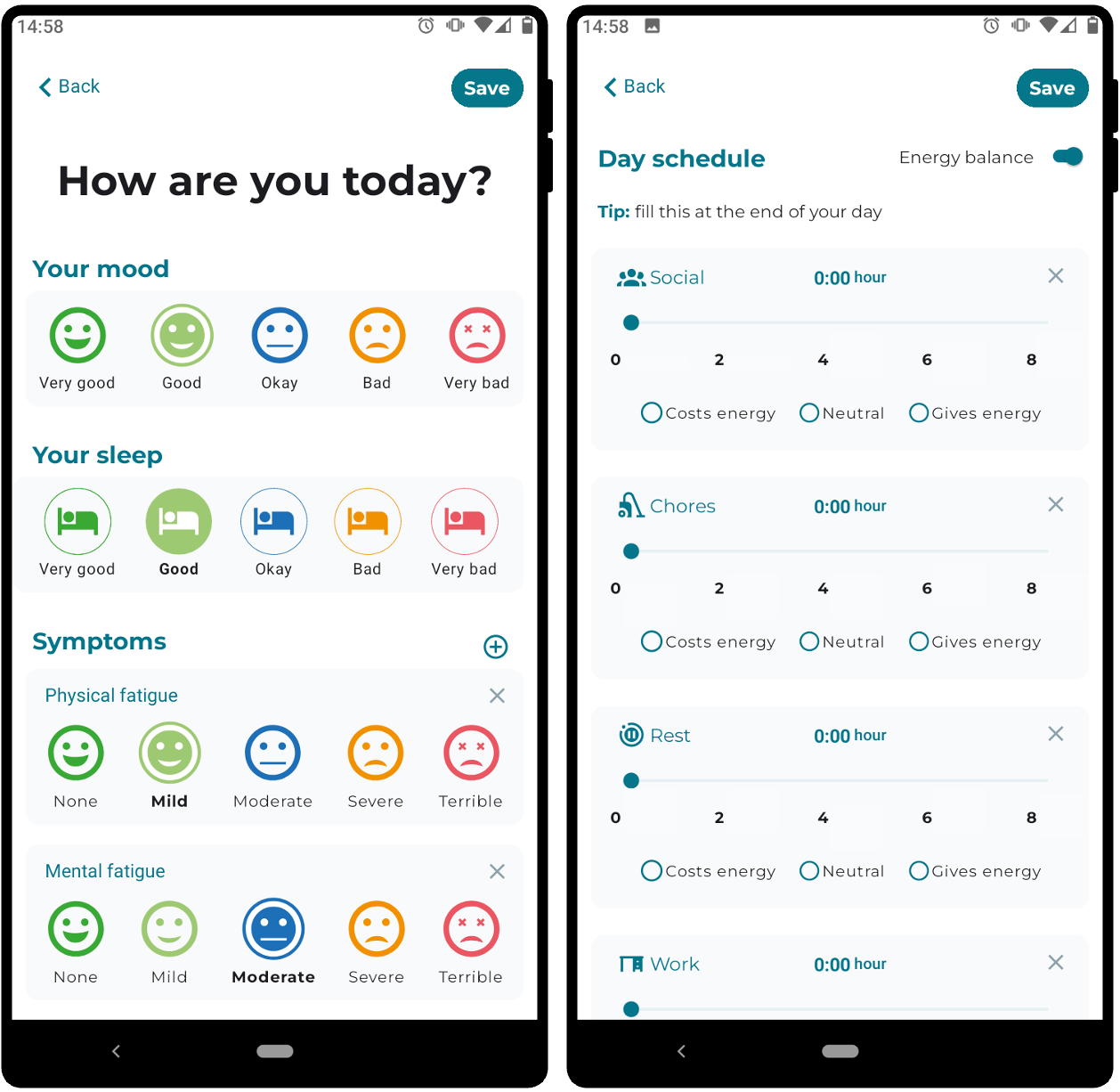}
    \caption{User interface of \texttt{HC}.}
    \label{fig:hc_ux_sample}
\end{figure}

\texttt{Healthy Chronos} is a digital diary for people living with the consequences of cancer (treatment), developed in cooperation with the Netherlands Cancer Institute and Dutch Cancer Society. The goal is to help this group cope with the physical and mental effects of dealing with cancer (treatment), though the app can accommodate any user-specified goal. Through reminders and surveys, the app helps users to score variables like mood, sleep and energy (scores on five-point scales), activities like work, social activities, chores, and exercise (in hours), and goals, like resting at least 30 minutes (booleans) (\autoref{fig:hc_ux_sample}). Users receive push messages to submit scores at the end of the day, and users can submit arbitrary queries about their data. \texttt{Healthy Chronos} has a free version and one free year of premium use through a waiver.\footnote{See \url{https://healthychronos.com/en/} for more information.}   

COACH was developed as an efficient Python pipeline to quickly generate counselling from user queries within the \texttt{Healthy Chronos} app. It was fully developed and evaluated in the Azure environment of \texttt{Healthy Chronos}, in line with the app's privacy and security agreements with users. COACH's full code and prompt is available on Github.\footnote{\url{https://github.com/yeem4n/llm-lifestyle-coach/}}

\begin{figure*}[t]
    \centering
    \includegraphics[width=1\linewidth]{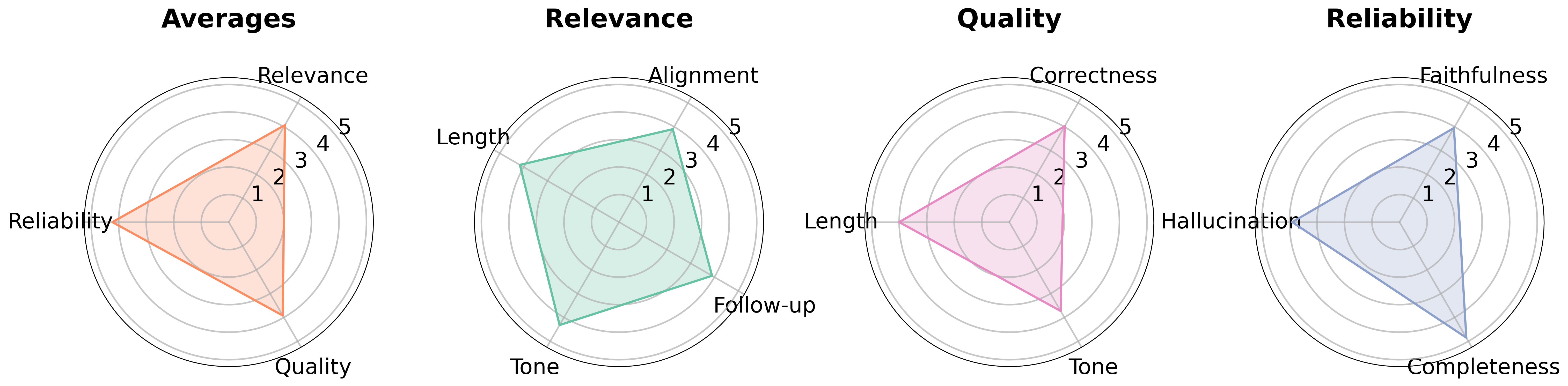}
    \caption{Visualization of QUORUM outcomes. `Averages' graph summarises relevance, quality, and reliability; `Relevance' and `Quality' summarize their respective dimensions by averaging over variable scores; `Reliability' contains rescaled variable ratios (\autoref{tab:eval_framework}.)}
    \label{fig:results_visualization}
\end{figure*}

COACH's workflow is as follows (also see \autoref{fig:pipeline_overview}). Once a single user query (or batch) is submitted to the application, COACH embeds the user query using the \textit{text-embedding-3-small} encoder from OpenAI. COACH then scrapes the subset of webpages from \texttt{kanker.nl} related to cancer and lifestyle, and converts them to a vector store so that knowledge stays up to date. Thereafter COACH retrieves the four chunks most relevant to the query. COACH transforms the table, and then appends it with chunks and query to a prompt. Prompt design followed the following insights from recent work:

\begin{itemize}
    \item Separating instructions and examples from `external' data, i.e., table representation and chunks from the knowledge database; it is known that keeping external information at the end improves performance \cite{sui_table_2024}; 
    \item Representing user tables in Markdown format, as LLMs are sensitive to different ways tabular inputs are serialized \cite{sui_table_2024}, and Markdown worked best for our use case (besides JSON and CSV).
    \item Providing descriptions of key column names and data types, e.g. ``headers containing `goal' always indicate booleans'', because headers are crucial to an LLM's capacity to understand column contents and relations to other columns \cite{singha2023tabular}.
    \item Drawing on in-context learning \cite{fang2024large}, by describing for three common topics in user queries (physical fatigue, mental fatigue, daily activities) the relevant columns in the user data. This `primes' the LLM to consider relevant columns and possible relations between them.
\end{itemize}

Current work in prompt engineering research has further shown that particularly in healthcare, a challenge is limited portability and scalability due to prompts becoming tied to specific domains, e.g. due to use of terminology \cite{wang2025prompt}. Hence we keep our approach straightforward, in that we leave specifics about the disease to the RAG component, and do not describe our medical population in detail. Our structuring and provision of examples in the prompt aims for portability, in that future users can easily adapt examples to their own data. In sum, our prompt is structured as follows: 

\begin{enumerate}
    \item A system message setting up the LLM as an AI assistant for addressing lifestyle-related questions for cancer patients;
    \item Instructions for obtaining an empathic tone, brevity in the response, and refraining from providing medical advice;
    \item A description of the user tables concerning headers (columns) and data types;
    \item Several examples of question topics and the appropriate columns to consider;
    \item Formatting instructions for separating user data analysis from counselling in the output;
    \item Concatenation of table, RAG chunks, and user query to instructions above. 
\end{enumerate}
We used LangChain, an open source framework to streamline LLM and RAG components \cite{chase_langchain_2022}, to build COACH.

The LLM used in the pipeline is GPT-4o (version 2024-11-20, accessed October 2025), due to the security and privacy requirements, and workflows already in place for \texttt{Healthy Chronos}. GPT-4o is currently still a good option in similar, recent systems and chatbots for lifestyle counselling \cite{jorke_2025,steenstra2024virtual}; suitably large open models could in future work take the place of GPT-4o.   

%MAG NIET The code of our pipeline is available at \footnote{\url{https://anonymous.4open.science/r/llm-lifestyle-coach/}}.  

\begin{table}[t]
\centering
\footnotesize
\setlength{\tabcolsep}{2pt}

\begin{minipage}[t]{0.31\columnwidth}
\vspace{0pt}
\centering
\begin{tabular}{@{}lc@{}}
\hline
\textbf{Var.} &  \\
\hline
Alignm. & $3.90_{\pm.94}$ \\
Fol. up & $3.90_{\pm1.15}$ \\
Tone & $4.32_{\pm.82}$ \\
Length & $4.16_{\pm.69}$ \\
\hline
\end{tabular}
\\[2pt]
Users
\end{minipage}\hspace{0.01\columnwidth}
\begin{minipage}[t]{0.31\columnwidth}
\vspace{0pt}
\centering
\begin{tabular}{@{}lc@{}}
\hline
\textbf{Var.} & \\
\hline
Correct & $4.02_{\pm.84}$ \\
Tone & $3.73_{\pm.75}$ \\
Length & $4_{\pm.57}$ \\
& \\
\hline
\end{tabular}
\\[2pt]
Experts
\end{minipage}\hspace{0.01\columnwidth}
\begin{minipage}[t]{0.31\columnwidth}
\vspace{0pt}
\centering
\begin{tabular}{@{}lc@{}}
\hline
\textbf{Var.} & \\
\hline
Faithfulness & $.79$ \\
Completeness & $.97$ \\
Hallucination & $.22$ \\
& \\
\hline
\end{tabular}
\\[2pt]
Developers
\end{minipage}

\caption{Main evaluation results. For users, we report averages over 19 users; for experts, averages over in total 100 evaluations (50 response-query pairs evaluated by two experts); for developers, ratios over in total 66 claims about user data (faithfulness/completeness) and 31 contextualisation statements (hallucination).}  
\label{tab:results}
\end{table}

\begin{table*}[t]
\centering
\footnotesize
\begin{tabular}{lcccccc}
\toprule
Annotator   & A (n=25)     & B (n=25)     & C (n=12)     & D (n=13)     & E (n=12)     & F (n=13)     \\
\midrule
Correctness & 4.08 (±0.63) & 4.88 (±0.33) & 3.33 (±0.47) & 4.15 (±0.36) & 2.75 (±0.73) & 3.92 (±0.47) \\
Tone        & 4.12 (±0.59) & 3.8 (±0.63)  & 3.42 (±0.76) & 3.7 (±0.92)  & 3 (±0.71)    & 3.85 (±0.36) \\
Length      & 4.16 (±0.46) & 4.04 (±0.34) & 3.92 (±0.28) & 4.46 (±0.63) & 3.25 (±0.6)  & 3.92 (±0.47) \\
\bottomrule
\end{tabular}
\caption{Mean and standard deviation of each expert and variable. N shows the number of query-response pairs evaluated per expert, each was labelled twice, totalling 100 observations.}
\label{tab:expert-annotation}
\end{table*}

\section{Results}\label{sec:results}
\subsection{User evaluation: Relevance}
Main results of applying QUORUM are given in \autoref{tab:results} and visually in \autoref{fig:results_visualization}. As can be seen, overall, users responded positively to the generated counselling. Results suggest that counselling aligned with personal circumstances, that tips were likely to be followed, and that the tone and length were satisfying, as all variable mean scores are around 4. However, we also see that variation in scores is larger for statements on alignment and follow-up, implying that users are more divided on this aspect than on tone and length. 

\paragraph{Remarks: Alignment} Inspection of user-submitted remarks provided more insight in what users appreciate in the counselling. Ten out of nineteen respondents submitted an additional remark, of which four respondents expressed that the response captured their situation very well, and was easy to understand. They also appreciated the level of detail in the answers, and the formatting that keeps claims about the data separated from the contextualisation (\autoref{fig:hc_ux_response_question}). In addition, two users emphasized that the advice felt more valuable due to the addition of specific references to their own diary data, which also helped them to better understand the contextualisation. %The references to user data support explainability and grounding, and as users can verify claims, they may also engender more trust and engagement. (speculatief)

\paragraph{Remarks: Misalignment} Four respondents expressed a misalignment in the counselling and their situation. One user, for example, reported the counselling recommended to seek help from friends and family, whereas for this user social activities were experienced as tiring, and as increasing burden for the social network, i.e., something to avoid. Other users mentioned a misalignment with their situation that is strictly speaking not related to the query-response pair, such as a lack of added functionality in app, or struggling with regularly submitting data at the end of the day. One user noted that the advice reiterated strategies they were already applying without positive outcomes, implying the generated response ideally takes into account previous counselling in the prompt context. 

Our user results suggest that the pipeline succeeds in generating advice that is relevant for users.

\subsection{Expert evaluation: Information quality}
Overall, experts were positive about the quality of generated counselling (\autoref{tab:results}, \autoref{fig:results_visualization}). They thought the counselling was often correct, and had mostly the right tone and length, with again most scores close to 4. but with smaller variation for each variable compared to users. Across 50 samples evaluated by 6 independent experts (minimal 2 evaluations per item with overlapping annotator pairs), Krippendorff's $\alpha$ was 0.602 for correctness. This indicates moderate agreement: the experts generally concur about answer correctness. Examining the per-expert mean scores, (\autoref{tab:expert-annotation}), no average scores for correctness were found closer to two or one, suggesting disagreements reflects differences in annotator strictness rather than strong divergence about system correctness. Thus, the pipeline rarely produces counselling that experts consider incorrect or harmful.

\paragraph{Remarks: Tone} % Bespaart ruimte
Regarding tone, experts diverged. Krippendorff’s $\alpha$ was 0.150 for tone and 0.272 for length, confirming that these dimensions were very subjective. Inspection of in total 79 remarks revealed that about a quarter raised concerns about tone and language use, describing the counselling as too condescending, direct, or unsympathetic. Also, the use of exclamation marks, imperatives, and motivational phrases were sometimes viewed negatively. Thus, experts see room for possible improvements, given that their evaluations were on average positive. 

%One risk here is that while the informational content of the answer might be often correct, if the communication style is not appropriate, users might be thrown off and less likely to consider the advice.  %Ik vind de antwoorden niet helder genoeg; zo staat er rustig opbouwen van routines, wat is rustig? Wat is dat positieve effect van rust, waar kan ik dan aan denken? Blijf goed luisteren naar je lichaam, hoe doe je dit, wat kan hierin bijdragen e.a. Niet voor iedereen is praten helpend, het is denk ik belangrijk om ook andere manieren te noemen zoals schrijven. Het is belangrijk om echt duidelijk te zijn, zodat dit mensen motiveert om deze adviezen op te volgen.

\paragraph{Remarks: Clarity}
Experts addressed clarity in about half of the remarks. Twenty-two remarks explicated that the counselling was not at a basic level (B1) of Dutch, hence should be rephrased. Experts considered some counselling statements on their own as too ambiguous, due to the lack of specific examples or further explanation, for example the phrase `listen to your own body' or `stay true to your routines', as it is not always clear what this entails. Experts emphasized that clarity and concreteness in lifestyle advice is important, as this renders counselling more actionable and memorable, in line with previous research \cite{szymanski_limitations_2025,degachi_towards_2025}. %Experts thus prioritized clarity and conciseness, rather than a level of detail that could potentially overwhelm the user. 

%Some experts noted that the same advice does not work for everyone, especially considering the subjective nature of experiences of certain activities. While for some patients, social activities might be very energy-draining, it could also have a positive effect on their mood. Ideally, the advice should be phrased with a bit more care and restraint. 

\paragraph{Remarks: Grounding} Five experts praised the explicit grounding of counselling in diary data, acknowledging the value of including insights from the user's diary. While some did note that care should be taken in not overgeneralizing over too little data, the references to concrete days, activities, or symptoms were seen as increasing transparency and trustworthiness. %This supports the use of structured personal data in a way that can promote user agency, and highlights the value of personalized health applications. 

Overall, though experts were overall quite positive about the counselling, they provided many remarks about how the counselling could be improved regarding its tone. In this respect they were more critical than users, a finding which we return to in the discussion.

%While the tips could be formulated more concretely and clearly in some cases, experts noted that the examples from the diary data , are a nice contribution and make the advice much more tangible. In this way, the user can verify for themselves what they experienced in the past, and reflect on their own diary data with added awareness and, ultimately, the applicability of the given advice. This ties into a larger notion of patient-centered care and patient autonomy. 

\subsection{Developer evaluation: Reliability}
Overall, we found that 79\% of claims about user patterns were faithful, i.e., consistent with the data. That is, about four out of five claims about user data are supported by the user's table. This also means that specific referenced dates, activities, or goals, from the user data are often correct, that is, concrete examples for patterns found in the user data. Further, regarding completeness, in 97\% of the claims (virtually all) the LLM succeeded in finding the relevant information, which often amounted to correctly retrieving relevant columns and their values to address a user query. Lastly, regarding hallucination, in 22\% of statements made in the contextualisation we found information not strictly present in the external knowledge database.

\begin{figure*}
    \centering
    \includegraphics[width=.95\linewidth]{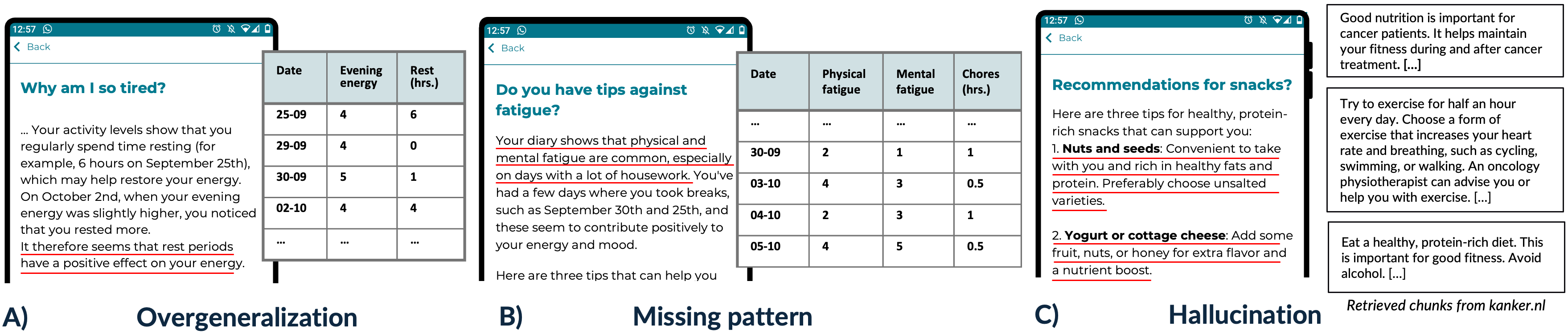}
    \caption{Examples of various errors in claims about data and contextualisation statements. A): overgeneralization of common-sense patterns: B) missing pattern in the user data; C): hallucination: advice is not strictly present in chunks.}
    \label{fig:errors_examples_researchereval}
\end{figure*}

\paragraph{Developer agreement}
Developer agreement was overall moderate across faithfulness, completeness, and hallucination (Cohen's $\kappa$ = 0.58), indicating some challenges in evaluating claims and statements. In particular cases with more ambiguous phrases, e.g. `\textit{The last \textbf{few} days...}' or `\textit{Scores are \textbf{often} between 3 and 5...}' led to differing judgments, despite establishing a list with definitions for ambiguous terms (provided in the Supplementary Materials). 

To better understand misalignment, we computed Positive Percentage Agreement for faithfulness, completeness and RAG context adherence, which were 85.39\%, 96\%, and 88.89\% respectively, while Negative Percentage Agreement was 48\%, 0\%, and 44.44\% respectively. This indicates that the two developers often agree whenever one developer evaluates a claim as faithful, complete, or hallucination-free, however, developers have much more difficulty on agreeing on the cases where according to one of them, the LLM was unreliable, incomplete, or provided information not present in the knowledge database.

\paragraph{Errors}
Though in most cases patterns in the data were extrapolated accurately, for faithfulness most errors arose from overgeneralisation, in which the model saw a pattern in the data that was not sufficiently strong or had also strong counterexamples. Example A in \autoref{fig:errors_examples_researchereval} illustrates this. Another issue was with missing patterns: sometimes COACH reported on a pattern that did not exist (example B in \autoref{fig:errors_examples_researchereval}). This could be seen as a type of hallucination, that in QUORUM is captured under faithfulness. 

Regarding hallucinations in contextualisation statements, we observed that the model sometimes goes beyond retrieved chunks in generating advice. Note that we had a strict interpretation of hallucination in \autoref{tab:eval_framework}: if a contextualisation statement suggested consuming healthy, protein-rich food, and the counselling contained examples (nuts, yoghurt, etc.) not strictly in the database, it was regarded an hallucination. Arguably, this could also be seen as a case of intuitive extrapolation rather than harmful content often associated with hallucinations. When such cases (example C in \autoref{fig:errors_examples_researchereval}) were not counted as hallucination, hallucination rate declined to 11\%. Remaining errors contained reporting of insights from user data, contrary to the instruction to the LLM in the prompt. We found no cases of hallucinations with wrong or harmful content. 

\section{Discussion \& Conclusion}
This study contributed QUORUM, a framework on evaluating relevance, quality and reliability of AI systems in health from three perspectives. The framework allows a future user to quickly glance how overall, and in what specific aspects, a given AI-driven system in health scores on topics important to different stakeholders (\autoref{fig:results_visualization}). As we have argued, a good evaluation of AI-driven systems in health should incorporate all perspectives in evaluation. We expect that AI-systems that evaluate on average positive on relevance, quality, and reliability, which in practice will mean around 4 or higher, will have higher chances of making positive real-world impact.

In our case, we saw convergence on a positive evaluation in the three perspectives, but also room for divergence. Divergence may reveal issues crucial for future development. In our case, we observed in expert remarks that they took issue with the clarity of generated counselling. Interestingly, this did not seem to resonate with scores or remarks of users, who indicated it was easy to follow. This signals our sample may be biased towards people with higher educational levels, as medical experts may have experience with a more varied population; this should inform the next iteration of COACH. 

Some divergence can also be found in sensitivity to errors. Alignment and correctness as variables are arguably for users and experts proxies for developers' more technical faithfulness and completeness metrics (though they do not fully overlap). Still, the scores and remarks of users and experts do not suggest grave errors in user-specific pattern extraction or hallucinated content from the database, though we know from the developer evaluation that roughly one in five claims is not consistent with user data or is not strictly traceable to the knowledge database \autoref{tab:results}. This could be because different stakeholders have their own backgrounds and biases that may impact their sensitivity to errors; QUORUM helps reveal such differences as starting point for further investigation.

We observed that if a user asks for example `Why am I so tired?', the retrieved chunks may contain biomedical details, for example on how changes in the immune system after cancer (therapy) can cause fatigue. Still, the counselling often defaults to more general tips, for example on sleep patterns instead of more biomedical facts. This could be an unwanted result of prompting the model to refrain from providing medical advice; yet, the point here is that neither users/experts scores or remarks seem to pick up on this. Especially experts could due to their training expect more comprehensive explanations in the counselling, that also includes more biomedical facts. A next iteration of COACH could focus on encouraging the LLM to include biomedical facts in the answer, which is not the same as providing medical advice.  

In conclusion, this paper demonstrated that relatively straightforward LLM-based systems like COACH can provide relevant, good quality, and reliable lifestyle counselling. We aimed to show that a multi-stakeholder lens is needed to evaluate the likelihood of real-world impact of any AI system in health. We expect that both QUORUM and COACH, as easy-to-understand framework and pipeline, will benefit other developers, experts, and patient population aiming to improve consumer health question answering with NLP systems. 

\section{Limitations}
There are several limitations that impact our work. While our work demonstrates the potential utility of our pipeline COACH and evaluation framework QUORUM to be extended to arbitrary populations and diseases, we only tested these in one context and use case. Future research should aim to test this approach in different contexts or use cases. In addition, our sample size of evaluated responses was rather limited due to time constraints, in particular on the hallucination rate in the contextualisation statements. Consequently, the observed alignment between hallucinated content and retrieved knowledge should be interpreted with caution. The LLM-generated counselling was also only based on the most recent <21 days of diary data. Future research could look into LLM performance on lifestyle counselling while considering patterns over longer periods of time in diary data. Finally, our pipeline only made use of one proprietary LLM, GPT-4o, for reasons addressed in \autoref{sec:methodolody:pipeline}. Future research should look into the use of open-source, ideally smaller, models. 

\section{Ethical Statement}
Our study protocol for recruitment of and data collection from users, medical experts, and developers was approved by the Leiden University Ethics Committee.  

\section{Acknowledgements}
We thank the users of \texttt{Healthy Chronos} who contributed as participants to this study. We are grateful to the cancer information specialists from \texttt{kanker.nl} for their time and effort in evaluating the responses. We also thank \texttt{Healthy Chronos} for the collaboration and access to the valuable platform. 

\section{Bibliographical References}\label{sec:reference}

\bibliographystyle{lrec2026-natbib}
\bibliography{lrec2026}

% \section{Language Resource References}
% \label{lr:ref}
% \bibliographystylelanguageresource{lrec2026-natbib}
% \bibliographylanguageresource{languageresource}

\end{document}